%% file: main.tex
\documentclass[letterpaper, 10 pt, conference]{ieeeconf}  
\IEEEoverridecommandlockouts                              %
\usepackage{00_ps}
\usepackage{arydshln}
\usepackage{subcaption}
\usepackage{graphicx}
\acsetup{patch/longtable=false}

\preprintfalse
\publishedtrue

\glsdisablehyper 
\input{00_acronyms.tex}
\input{00_copyright.tex}

\def\BibTeX{{\rm B\kern-.05em{\sc i\kern-.025em b}\kern-.08em
  T\kern-.1667em\lower.7ex\hbox{E}\kern-.125emX}}

\begin{document}
\title{\LARGE \bf
    CoMoCAVs: Cohesive Decision-Guided Motion Planning for Connected and Autonomous Vehicles with Multi-Policy Reinforcement Learning 
    \thanks{This work is based on my Master’s thesis, conducted under the supervision of Jianye Xu (RWTH Aachen University, Germany). It presents preliminary results from our ongoing work, and we may revise or extend the findings in the future. This work is under preparation for publication. Computations of this work were performed with computing resources granted by RWTH Aachen University under project thes1898.}
}

\author{
    Pan Hu$^{1}$\,\orcidlink{0009-0004-6683-8778}
    \thanks{$^{1}$Department of Computer Science, RWTH Aachen University, Germany, \texttt{pan.hu@rwth-aachen.de}}
}
    \maketitle
\thispagestyle{IEEEtitlepagestyle}
\begin{abstract}

Autonomous driving requires reliable and efficient solutions to tightly interrelated problems such as decision-making and motion planning. In our context, decision-making specifically refers to highway lane-selection and motion planning is about generating control commands (speed and steering) to reach the chosen lane. In the context of Connected Autonomous Vehicles (CAVs), achieving flexible and safe lane selection decisions alongside precise trajectory execution remains a significant challenge. This paper proposes a novel framework—Cohesive Decision-Guided Motion Planning (CDGMP)—that tightly integrates these two modules using a Mixture of Experts (MoE) inspired architecture combined with a multi-policy reinforcement learning (MPRL) approach. By leveraging multiple specialized sub-networks coordinated through a gating mechanism, our method decomposes the complex decision-control task into modular components, each focusing on a specific aspect of the driving task. This design not only improves computational efficiency by activating only the most relevant sub-policies during inference but also maintains safety guarantees through modular specialization.

CDGMP enhances the adaptability and robustness of CAVs across diverse traffic scenarios, offering a scalable solution to real-world autonomy challenges. Moreover, the architectural principles behind CDGMP—particularly the use of MoE—provide a strong foundation for extending the framework to other decision-making and control problems in high-dimensional domains. Simulation results (available at \url{https://youtu.be/_-4OXNHV0UY}) validate the effectiveness of our framework, demonstrating its ability to deliver reliable performance in both lane selection and downstream motion planning tasks.

\par\medskip
\end{abstract}

\section{Introduction}\label{sec:introduction}

Connected and autonomous vehicles (CAVs) is a hot topic in both academia and industry, largely due to its potential of automating humanity's most ubiquitous activity—transportation. This field has witnessed an architectural evolution: from traditional rule-based modular systems, to hybrid supervised learning systems with rule-based fallbacks, and more recently toward end-to-end (E2E) learning paradigms, by leveraging a large neural network learn from vast amounts of human driving experience, directly translating sensor inputs into vehicle control commands. However, current research and industrial applications predominantly employ supervised learning-based modular architectures due to computational and data constraints. Our work builds upon this prevalent framework, which typically consists of four core modules: (1) Perception, which interprets sensor data to identify objects and road features; (2) Decision Making, which determines high-level maneuvers like lane changes or stops; (3) Motion Planning, which generates collision-free trajectories; and (4) Control, which executes these trajectories through precise actuation commands\cite{machines5010006}.

While supervised deep learning approaches for autonomous driving rely heavily on costly expert-labeled datasets, reinforcement learning (RL) has emerged in recent years as a transformative alternative for training CAVs. RL's reward-driven paradigm enables CAVs to learn optimal behaviors through direct environmental interaction, eliminating the need for pre-collected labeled data\cite{INAMDAR2024100810}. This iterative learning process not only enhances generalization across diverse driving scenarios but also proves more scalable and cost-effective than supervised methods, as the system autonomously refines its policies through continuous experience.

Our work specifically targets highway lane-changing scenarios by introducing a novel decision-making and motion planning framework for CAVs. By leveraging multi-policy reinforcement learning (MPRL), our solution tightly couples lane selection decisions with motion planning. This cohesive integration, where lane-change decisions immediately trigger corresponding motion policies, equips CAVs with sufficient flexibility to handle road emergencies, as both modules operate synchronously under an MPRL coordination mechanism. Our proposed framework, inspired by mixture of experts (MoE) mindset, is in general able to solve complex problems with multiple small networks.

\subsection{Related Work}
\label{sec:related}

Decision-making in autonomous driving determines high-level strategic actions (e.g., stopping, lane-keeping, or lane changes) without specifying low-level control parameters~\cite{garrido22}. This abstraction layer focuses exclusively on maneuver selection based on real-time traffic conditions and route requirements, while delegating execution details to downstream controllers. The separation of strategic decision-making from operational control enables modular system architecture and verifiable safety guarantees.

Supervised learning approaches using explicitly labeled datasets currently dominate research in this domain, as evidenced by recent comprehensive surveys\cite{li2018survey}\cite{qi2021dmsurvey}. However, rule-based systems remain widely deployed in production vehicles due to their robustness and interpretability. Prior work has demonstrated their particular effectiveness in complex scenarios like mixed-traffic intersections, where deterministic decision logic offers advantages in safety verification and regulatory compliance\cite{aksjonov2021rulebasedad}.

Our work specifically focuses on lane change decision-making, where RL plays a significant role in this research domain. Open-source RL simulation platforms designed for lane change scenarios, such as SUMO, HighwayEnv, and DeepTraffic, have been widely adopted in relevant studies\cite{WANG2025105005}\cite{highway-env}\cite{fridman2019deeptrafficcrowdsourcedhyperparametertuning}. These approaches and many alike ones share a common simplification: they assume an instantaneous lane shift is realized at the next timestep after a decision is made, while ignoring the motion control process that gradually moves the vehicle towards the target lane\cite{capello2023}\cite{bellotti2023}. This abstraction intentionally ignores motion control aspects to concentrate exclusively on lane change decision-making, trading realism for experimental convenience. In short, these works focus solely on making correct lane-selection decisions, without the additional — yet essential — consideration of the motion planning component.

Several studies have increased system complexity by incorporating motion planning, yet they typically treat lane-changing as an atomic operation that cannot be aborted during execution. Similar to this work, they often employ a hierarchical architecture where decision-making and motion control operate at different frequencies (e.g., \SI{1}{\hertz} vs \SI{50}{\hertz}), rendering decisions effectively irrevocable once initiated\cite{han2022multiagentreinforcementlearningapproach}.

Notably, the random exploration inherent in RL algorithms may lead to insufficient high-quality training data, potentially compromising the safety of learned policies. This limitation highlights the importance of incorporating imitation learning from human demonstration data to improve both the safety and practicality of RL-based decision strategies\cite{wu2022safedecisionmakinglanechangeautonomous}.

To address the ``irrevocable'' decision-making issue, recent works have proposed synchronized decision-making and motion planning approaches. Many studies in this combined research direction employ hierarchical reinforcement learning (HRL) framework. Work~\cite{pateria2021} conducts a comprehensive survey of HRL applications in this domain. Work~\cite{naveed2020trajectoryplanningautonomousvehicles} introduces an RL system featuring two policy networks operating in a semi-parallel architecture. The first network handles lane change decisions while the second works with a trajectory planner and PID controller to execute these decisions. The system demonstrates improved coordination between decision-making and execution layers. An alternative hierarchical structure is proposed, operating sequentially rather than in parallel\cite{lu2023actiontrajectoryplanningurban}. The initial policy generates decisions, followed by a secondary policy that produces trajectories based on these decisions, with PID control handling final execution. 

While these approaches represent progress, they share a common limitation: their motion planning relies on selecting predefined waypoints along the roadway as destination, which is inadequate to capture real-world driving dynamics, as in~\cite{han2022multiagentreinforcementlearningapproach}. For instance, work~\cite{sadat2019jointlylearnablebehaviortrajectory} develops a motion planning-oriented decision-making solution, using the quality of selected trajectory dots as key evaluation metric for the whole system. To address the ``impractica'' limitation, this comprehensive modular system integrates perception, decision-making, motion planning and execution components, greatly simulated industrial practice~\cite{li2022combinetrajdec}. However, it still remain constrained by their reliance on model-based methods for path generation and motion control, limiting their generalizability to novel roadway configurations.

\subsection{Contribution}

Our work proposes a novel framework for decision-making and motion planning in CAVs that employs tightly coupled decision-motion integration. High-level decisions (e.g., lane changes) synchronously trigger corresponding low-level motion policies to execute optimized trajectories in real time. Inspired by MoE  mindset and fused with a divide-and-conquer approach, the framework decomposes complex tasks into specialized policy networks, each handling a specific scenario. Highlights of our proposed framework include:

\begin{itemize}
    \item \textbf{Scalability:} The MoE-inspired MPRL design allows incremental addition of policies for new scenarios (e.g., intersections, merges) without retraining the entire system;
    \item \textbf{Interpretability:} Specialized networks simplify debugging and auditing compared to monolithic black-box models;
    \item \textbf{Flexibility:} Decision-making and motion planning are executed sequentially at every timestep, and each decision can be revised at the next timestep;
    \item \textbf{Efficiency:} At a given timestep, only one of these specialized small networks is activated to execute motion control, achieving greater efficiency compared to monolithic models.
\end{itemize}

\subsection{Structure}

Section~\ref{sec:prel} introduces the fundamental concepts relevant to this work, presenting both standard definitions and our specific interpretations. Section~\ref{sec:propm} details the proposed methodology, including its formal mathematical formulations. Section~\ref{sec:exp} describes the experimental environment configuration, training methodology, and presents corresponding simulation results validating our approach. Finally, Section~\ref{sec:conclusions} summarizes our key contributions and identifies promising directions for future research and optimization.




    

\section{Preliminary}\label{sec:prel}
\subsection{Decision-Making in CAVs}
Decision-making in CAVs represents the core intelligence that interprets environmental inputs and determines safe, efficient driving strategies. Unlike traditional rule-based approaches, modern CAV systems employ data-driven methods—such as RL and probabilistic reasoning—to dynamically assess traffic conditions, predict agent behaviors, and execute maneuvers (e.g., lane changes, merges, or emergency stops). This capability hinges on real-time coordination between perception, planning, and control modules, ensuring adaptability in complex, uncertain environments while adhering to safety and regulatory constraints. 

Our study specifically addresses highway lane-change decision-making. Consider an ego-vehicle navigating a highway system with $n$ parallel lanes sharing the same heading direction. Our high-level decision policy $\mathrm{P}_{\mathrm{HL}}$ takes the current state $s_{t}$ as input and outputs a discrete decision $o_{HL} \in \{0,1,2,...,n-1\}$ , specifying the target lane for timestep $t+1$, where $t$ refers to current timestep.

\subsection{Motion Planning in CAVs}

Motion planning in CAVs serves as the computational bridge between high-level decisions and low-level vehicle control, transforming strategic driving intentions into safe, dynamically feasible trajectories. Operating within a continuous state space, motion planners account for vehicle dynamics, obstacle avoidance, passenger comfort, the interaction with other agents in the environment and traffic rules to generate smooth, collision-free paths. 

Under our context, motion planning specifically refers to generating vehicle control commands: $v$ (speed) and $\delta$ (steering angle), aims to track a given reference path, specifically the centerline of the target lanelet. The objective is to ensure the ego-vehicle follows this reference path precisely while optimizing for driving efficiency and guaranteeing collision-free trajectories. 

\subsection{Reinforcement Learning}

RL is an interdisciplinary area of machine learning and optimal control concerned with how an intelligent agent should take actions in a dynamic environment in order to maximize a reward signal. Reinforcement learning is one of the three basic machine learning paradigms, alongside supervised learning and unsupervised learning\cite{wikirl}.

Technically, RL is about having an agent interacts with an environment by observing states $s_{t}\in \mathcal{S}$, where $t$ refers to the current timestep, taking actions $a_{t}\in \mathcal{A}$ according to a policy $\pi(a\mid s)$, and receiving rewards $r_{t} \sim \mathcal{R}(s_{t}, a_{t})$ while transitioning to new states $s_{t+1}\sim \mathcal{P}(s_{t+1}\mid s_{t}, a_{t})$, with the goal of maximizing the expected discounted return $\mathbb{E}\left[\sum_{k=0}^{\infty} \gamma^{k} r_{t+k}\right]$ where $\gamma \in [0,1)$ is the discount factor, smaller $\gamma$ means more focus on short-term reward, higher $\gamma$ means the system values long-term reward more\cite{introrl}. For example, an autonomous vehicle agent learns optimal driving behaviors by receiving positive rewards for smooth lane-keeping and negative rewards for collisions, continually refining its policy through this reward-driven environmental interaction.

Multi-Agent Reinforcement Learning (MARL) is a subfield of reinforcement learning that studies the behavior of multiple learning agents coexisting in a shared environment \cite{marl-book}. Typically, each agent aims to maximize its own reward, and the training paradigm is categorized as either centralized (shared observations) or decentralized (agent-specific observations). While centralized training can foster potential collaboration among agents\cite{liao2023}, our work adopts decentralized training to enhance experience collection and accelerate convergence.

MPRL refers to reinforcement learning systems that involve multiple policy networks. While no formal definition of this term yet exists, the prevailing approach in MPRL involves training distinct policies either for multiple objectives or to enhance system robustness\cite{kume2017mapbasedmultipolicyreinforcementlearning}\cite{DING2024101550}. In our work, we specifically employ multiple policies trained to achieve different objectives within the learning framework.

\section{Proposed Method}
\label{sec:propm}

\begin{figure}[t!]
    \centering
    \includegraphics[width=1.0\linewidth]{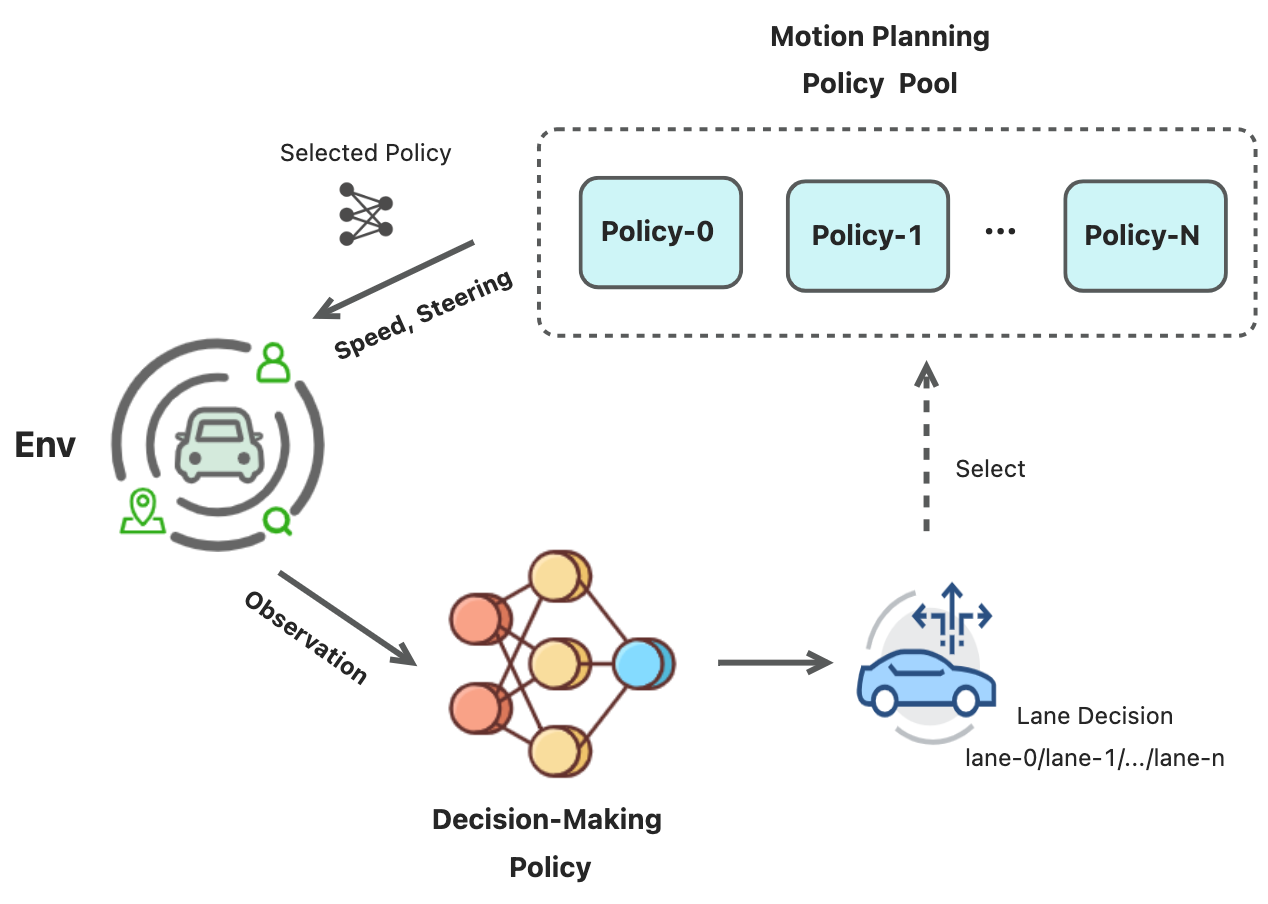}
    \caption{Illustration of the proposed framework.}
    \label{fig:proposed_archi}
\end{figure}

The proposed framework has a novel architecture that tightly integrates decision-making and motion planning of CAVs through the MoE strategy. This architecture is composed of two layers, the high-level layer is a decision-making module and the subsequent low-level layer is the motion planning module. This kind of hierarchical design enables efficient coordination between strategic decision layers and motion execution modules, offering a practical solution for real-world CAV deployment in dynamic environments. Coordinated within a MPRL framework, the two modules ensure real-time responsiveness while maintaining strategic consistency with the vehicle's global route objectives. An illustration of our proposed architecture is shown in Figure~\ref{fig:proposed_archi}.

\subsection{High-Level Module}

\subsubsection{Backbone Network}
The high-level module bears the role of decision-making, consists of a single neural network. In our experiments, we evaluated and compared the performance of multiple standard deep neural networks, including MLP, LSTM, and encoder-only Transformer\cite{lstm}\cite{vaswani2023attentionneed}. These backbone networks are substitutable through system initiation arguments.

\subsubsection{Observation Schema and Module Outputs}
The high-level module takes vector-based bird-eye view (BEV) observation of the environment as input and outputs a target lane selection. The observation vector for the high-level policy network has a fixed length of 33. Its structure is formally defined in Table~\ref{tab:hl_ob}.

\begin{table}[htbp]
\centering
\caption{Observation of high-level module.}
\begin{tabular}{{ll}}
\hline
\textbf{Terms} & \textbf{Object} \\ 
\hline
Position, rotation, speed, lane & Ego-vehicle \\ 
\\
Position, rotation, speed, lane & Surrounding vehicles \\
Distance\;to\;ego-vehicle  &    \\
Is\;ahead\;of\;ego-vehicle   &  \\
\hline
\end{tabular}
\label{tab:hl_ob}
\end{table}

The output from high-level module is an one-hot vector, the length of the vector is all the available choices, in our work, it is the total number of car lanes on the driveway. The action space for high level policy is $\mathcal{A}_{\mathrm{HL}}=\{0,1,2,3…n-1\}$, where $n$ stands for the total car lanes available in the environment. In our simulation environment, there are two lanes available, hence the $n$ equals 2, the discrete action space is specifically $\mathcal{A}_{\mathrm{HL}}=\{0,1\}$, $0$ represents for driving on the lane-0 (left lane) and $1$ for lane-1 (right lane). The decision made by the high level policy at timestep $t$ is marked as $decision_{t}$, for instance, $decision_{t}=0$ stands for driving on the left lane on timestep $t$.

\subsubsection{Rewarding Schema}
The high-level policy's reward mechanism, as presented in Table~\ref{tab:rwd_hl}, is designed to optimize lane selection decisions by balancing three critical objectives: (1) minimizing unnecessary lane changes to maintain driving consistency, (2) enforcing strict safety requirements through severe collision penalties, and (3) proactively avoiding lanes with slower-moving vehicles ahead. The primary objective of the high-level decision-making module is to navigate the vehicle into a lane that is free of slower-moving traffic ahead, thereby reducing the risk of potential collisions and enhancing overall driving efficiency. When the ego-vehicle approaches a slow-moving vehicle ahead of it in the same lane, and the distance between these two is within the pre-defined safety threshold, a penalty is applied in proportion to their distance. The closer the ego-vehicle to the slow-moving vehicle is, the higher the penalty incurred. Additionally, to discourage unnecessary lane changes, a small penalty is assigned for every lane change, regardless of the surrounding traffic conditions.

\begin{table}[htbp]
\centering
\caption{Rewarding schema for high-level module.}
\begin{tabular}{{lll}}
\hline
\textbf{Terms} & \textbf{Type} & \textbf{Value} \\ \hline
Collision with other agents & Penalty & Fixed \\
Lane change    &   Penalty & Fixed \\
Driving on risky lane  &   Penalty & Proportional \\
Driving on safe lane   &   Reward &  Fixed \\
\hline
\end{tabular}
\label{tab:rwd_hl}
\end{table}

\subsection{Low-Level Module}

\subsubsection{Introduction}
The low-level module comprises a pool of specialized expert policies, where each policy is specifically trained to handle a distinct driving scenario. In our simulation setup, we implemented two such policies: one optimized for navigating to and then keep driving on left lane and another for right lane, all of these policy networks are simple MLP. The training process for these networks involves scenario-specific conditioning, for instance, when training the left-lane driving policy, vehicles are randomly initialized across both lanes of our highway simulation environment while having their long-term and short-term reference paths permanently set to the left lane. Through extensive training, each policy network successfully learns to navigate from any initial position on the map to its designated target lane then keep driving along the target lane's centerline.

\subsubsection{Observation Schema and Module Outputs}

The observation vector for low-level module has a length of 57, as presented in Table~\ref{tab:ll_obs}, contains more fine-grained details comparing to the observation schema used in high-level module.

\begin{table}[htbp]
\centering
\caption{Observation of low-level module.}
\begin{tabular}{{ll}}
\hline
\textbf{Terms} & \textbf{Object} \\ 
\hline
Position, rotation, velocity & Ego-vehicle \\
Short-term reference path & \\
Distance to its reference path & \\
Nearby lane boundaries & \\
\\
Position, rotation, speed, lane & Surrounding vehicles \\
Distances to ego-vehicle  &  \\
Short-term reference path   & \\
\hline
\end{tabular}
\label{tab:ll_obs}
\end{table}

The low-level policy's action space is defined as $\mathcal{A}_{\mathrm{LL}}=(a_{v},a_{\delta})$ , $a_{v}$ stands for speed and $a_{\delta}$ stands for steering. Since low-level module contains multiple specialized policies, these parameters are policy-specific: $a_{v}\in\{a_{v,l_{0}},a_{v,l_{1}}\}$ and $a_{\delta}\in\{a_{\delta,l_{0}},a_{\delta,l_{1}}\}$, with $l_{0}$ and $l_{1}$ denoting parameters from the lane-0 and lane-1 policies respectively. At timestep $t$, $a^{t}_{v,l_{0}}$ specifically indicates the speed parameter generated by the lane-0 policy. Both control parameters are constrained to physically feasible intervals to respect vehicle dynamics: $a_{v}\in[v_{min},\;v_{max}]$ for speed and $a_{\delta}\in[\delta_{min},\;\delta_{max}]$ for steering.

\subsubsection{Rewarding Schema}

In order for the policy network to effectively learn both target lane reaching and centerline tracking, a carefully designed reward function plays a vital role. The reward schema for the low-level module incorporates multiple critical factors: (1) reaching target lane, (2) precise centerline adherence after reaching the target lane, (3) collision avoidance. All the rewarding terms are presented in Table~\ref{tab:rwd_ll}.

\begin{table}[htbp]
\centering
\caption{Rewarding schema for low-level module.}
\begin{tabular}{{lll}}
\hline
\textbf{Terms} & \textbf{Type} & \textbf{Value} 
\\ \hline
Collision with lane boundaries  &  Penalty & Fixed \\
Collision with other agents   &   Penalty &  Fixed \\
Too close to other agents  &  Penalty  & Proportional \\
Deviating from lane centerline  &  Penalty & Proportional \\
Deviating from reference-path  &  Penalty & Proportional \\
Steering too quick  &   Penalty & Proportional \\
Forward movement  &   Reward & Proportional \\
High velocity   &   Reward & Proportional \\
\hline
\end{tabular}
\label{tab:rwd_ll}
\end{table}

\section{Experiment}
\label{sec:exp}

\subsection{Environmental Setup}

\begin{figure}[t!]
    \centering
    \includegraphics[scale=0.8]{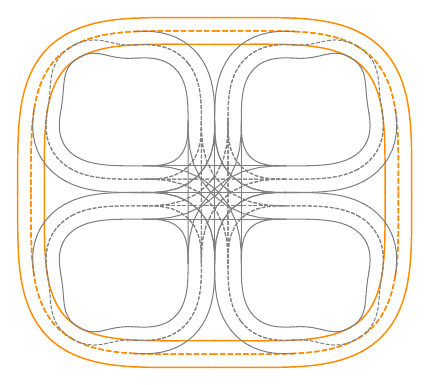} 
    \caption{CPM Lab map.}
    \label{fig:cpm_map}
\end{figure}

All experiments are conducted using the map from CPM Lab~\cite{cpm_lab} and largely adhering to the vector-based environmental representations proposed in SigmaRL framework~\cite{xu2024sigmarl}. To accurately replicate highway driving conditions, our implementation specifically utilizes the two outer circular lanes of the CPM Lab map (marked in orange in Figure~\ref{fig:cpm_map}), effectively creating a standardized two-lane highway environment. This configuration provides a controlled yet realistic setting for evaluating lane-changing behaviors while maintaining consistent road geometry throughout testing.

The simulation environment incorporates intentionally designed risky scenarios through the strategic placement of slow-moving vehicles randomly distributed across both lanes. These slow vehicles, programmed to maintain consistent lane discipline at significantly reduced speeds, create challenging situations where the well-trained high-level policy must identify and select optimal lanes free from obstructing traffic. The system's configurability allows adjustment of both the number of autonomous agents and slow vehicles within each environment, enabling scalable testing of the decision-making policy's ability to navigate around traffic obstacles while maintaining safe and efficient driving trajectories.

Low-level motion planning is guided by a lanelet-based~\cite{lanelet} reference path mechanism. The map is composed of hundreds of lanelets, with the short-term reference path always aligned to the centerline of the current lanelet. For instance, if at time step $t$, the ego-vehicle is in the left lane and the decision for $t+1$ is to change to the right lane, the reference path is updated to the centerline of the nearest lanelet in the right lane. This mechanism offers clear lane-following guidance and encodes each vehicle’s decision intention, enriching the representation of both the road environment and surrounding traffic participants.

\subsection{Training Setup}

The training proceeds under MARL setup, which accelerates convergence by enabling simultaneous experience collection from multiple agents within each episode. This approach significantly improves sample efficiency compared to traditional single-agent methods. During training, we use an experience replay buffer mechanism: the environment runs for a full episode with a predefined maximum steps, after which all collected frames are stored in the buffer. Each training iteration processes a mini-batch sampled from this buffer, with multiple epoch updates applied before proceeding to the next episode. The core hyperparameters governing the training is listed in Table~\ref{tab:hyps}. In which, ``lr'' refers to the initial learning rate, it gradudally reduces to ``lr\_min'' during training.

\begin{table}[htbp]
\centering
\caption{Key hyper-parameters.}
\begin{tabular}{{ll}}
\hline
Hyper-parameter & Value \\
\hline
n\_iters & 200 \\
frames\_per\_batch & 4096 \\
num\_epochs & 70 \\
minibatch\_size & 512 \\
lr & 1e-4 \\
lr\_min & 1e-5 \\
max\_grad\_norm & 1.0 \\
clip\_epsilon & 0.1 \\
gamma & 0.99 \\
lambda & 0.9 \\
entropy\_epsilon & 0.1 \\
max\_steps & 512 \\
\hline
\end{tabular}
\label{tab:hyps}
\end{table}

Furthermore, the low-level and high-level modules are trained sequentially and independently. First, all these low-level experts are trained to control the vehicle to reach to and keep driving on the designated lanes, wherever is the ego-vehicle’s initial position. After these low-level skills are mastered, the high-level module is then trained to make decent lane decisions for safety, dynamically selecting and deploying the appropriate low-level experts to execute the selected lane selection decisions. All of the policies in high-level and low-level modules are trained with PPO algorithm.

\subsection{Evaluation Metric}

Slow-moving obstacle vehicles are distributed across both driving lanes, requiring the ego-vehicle to consecutively make correct lane selection decisions in the challenging area of the loop to avoid collisions with these slow vehicles; thus, the decision correctness serves as our core evaluation metric. To align with our simulation setup, the final metric quantifies the total collision count across all agents over 2,000 simulation steps (approximately 3 minutes), directly measuring the framework’s success in collision avoidance. In our comprehensive experiments, we compared the performance of different network architectures, including Transformer, MLP, and LSTM. We also evaluated the impact of varying safety thresholds. 

To demonstrate the robustness of our proposed method, we conducted simulations using six different environment seeds, effectively diversifying the evaluation set in a manner analogous to dataset variation in supervised learning. We then compared both the average and best performance achieved under each seed. 

\subsection{Results and Discussion}

\subsubsection{Impact of Backbone Networks}

Our experiments comprehensively evaluated the collision avoidance performance of the proposed framework by comparing it against implementations of classical backbone networks in their naive forms. These comparisons demonstrate both the efficiency and generalization capability of our approach, showing that even minimally optimized networks can effectively solve this complex problem when integrated with our framework.

\begin{figure}[htbp]
    \centering
    \includegraphics[width=0.8\linewidth]{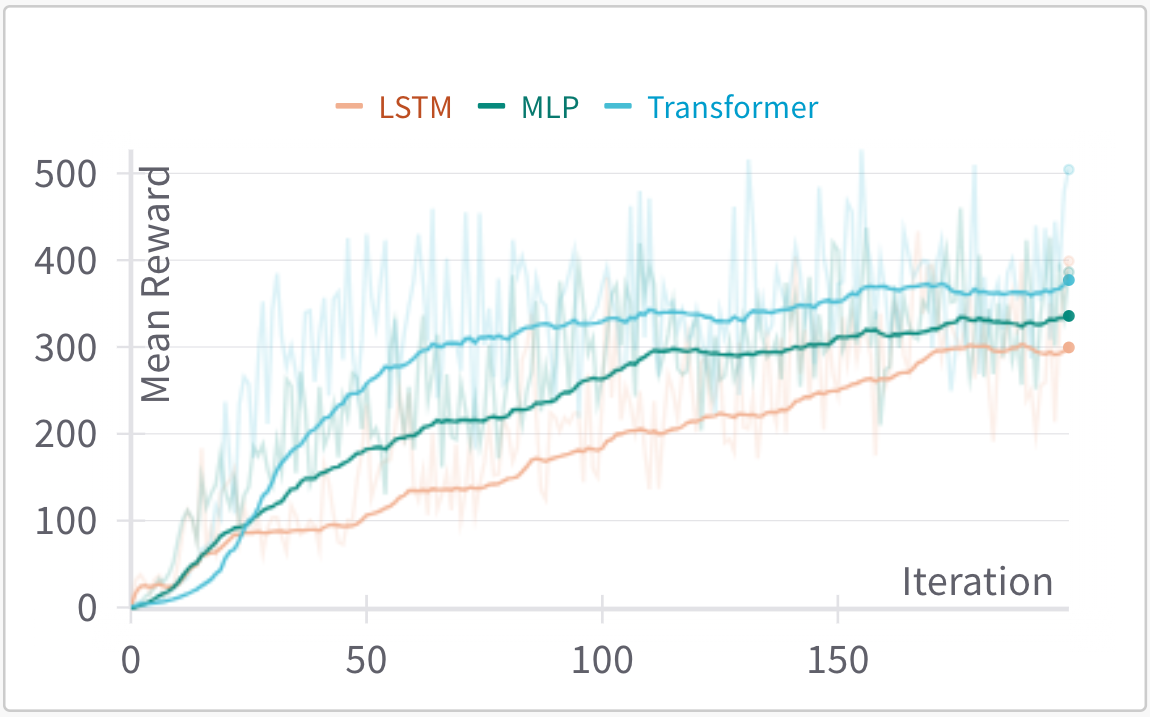}
    \caption{Episode reward from training.}
    \label{fig:episode_reward}
\end{figure}

As shown in Figure~\ref{fig:episode_reward}, the Transformer demonstrates superior convergence capability, achieves the highest reward score. As presented in Table~\ref{tab:performance}, the evaluation result shares the same trend as the training reward curve. This indicates that our reward scheme is effectively aligned with our ultimate objective.

As shown in Table~\ref{tab:performance}, the Transformer model demonstrates superior collision avoidance performance. It averages 6.67 collisions over 2,000 simulation steps across six different environment seeds, with the best performance reaching as low as 2 collisions. This clearly outperforms both the LSTM and MLP models, which average 7.17 and 8.17 collisions, respectively.

\begin{table}[htbp]
\centering
\caption{Collisions over backbone networks.}
\begin{tabular}{{llll}}
\hline
\textbf{Backbone Network} & \textbf{Avg.} & \textbf{Std.} & \textbf{Best}\\ 
\hline
\underline{Transformer}  &  \underline{6.67} & \underline{1.64} & \underline{2} \\
LSTM   &  7.17 & 2.32 & 5  \\
MLP  &  8.17 & 2.32 & 6 \\
\hline
\end{tabular}
\label{tab:performance}
\end{table}

\begin{figure}[t]
    \centering
    \begin{subfigure}[t]{0.45\linewidth}
        \centering
        \includegraphics[width=\textwidth]{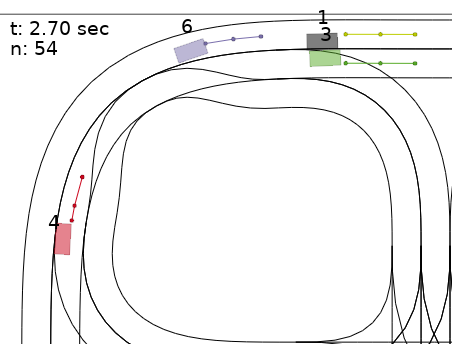}
        \caption{Timestep 54.}
        \label{fig:lc_54}
    \end{subfigure}
    \hfill
    \begin{subfigure}[t]{0.45\linewidth}
        \centering
        \includegraphics[width=\textwidth]{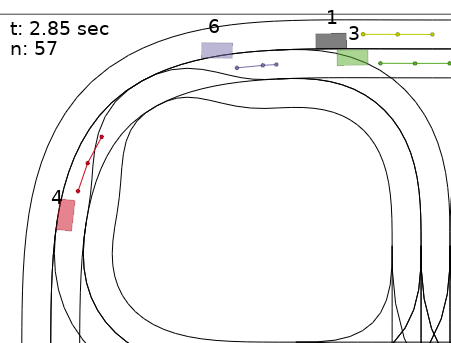}
        \caption{Timestep 57.}
        \label{fig:lc_57}
    \end{subfigure}
    
    \vspace{0.5em}
    
    \begin{subfigure}[t]{0.45\linewidth}
        \centering
        \includegraphics[width=\textwidth]{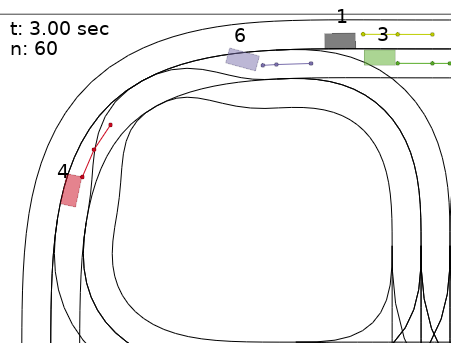}
        \caption{Timestep 60.}
        \label{fig:lc_60}
    \end{subfigure}
    \hfill
    \begin{subfigure}[t]{0.45\linewidth}
        \centering
        \includegraphics[width=\textwidth]{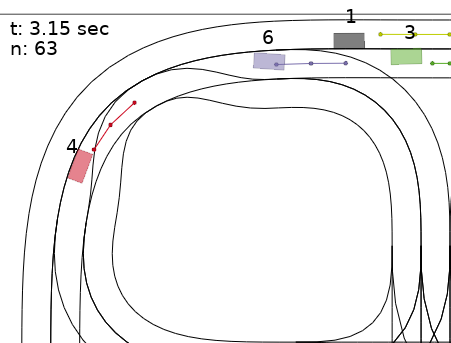}
        \caption{Timestep 63.}
        \label{fig:lc_63}
    \end{subfigure}
    
    \caption{
        Lane changing scenario.
    }\label{fig:sim_frames}
\end{figure}

The snapshots shown in Figure~\ref{fig:sim_frames} illustrate how the CAVs (depicted in non-gray colors) are controlled by our RL framework to make correct lane-selection decisions and avoid collisions with slower vehicles (shown in gray). The colored, connected dots are the lanelet-based reference paths of each CAV, it shows the currently planned target lanelet. In particular, the purple CAV (numbered 6) has a slow vehicle (number 1) directly ahead on its current lane. To maintain high speed and avoid a potential collision, it must change to the right lane. Our RL framework evaluates the current state, including the positions of surrounding traffic participants and the distance to the slow vehicle, to determine an appropriate time to initiate the lane change maneuver.

Each successful lane change requires the CAV to make around 10 consecutive correct decisions to reach a stable driving state in the target lane. This property, referred to as decision consistency, highlights the robustness of our proposed method. On one hand, our system can persist with a correct decision over an extended period; on the other hand, decisions can be revised at every next timestep. This reflects one of the core principles of our approach: flexible, yet reliable, decision-making.

\subsubsection{Impact of Safety Threshold}
Table~\ref{tab:performance_thres} illustrates the impact of different safety threshold settings on performance. Setting this threshold too high or too low can both negatively affect collision avoidance. In our reward schema, all lane change decisions are penalized. Additionally, we aim to minimize lane changes, as excessive switching can diminish passenger comfort and increase danger in real-world scenarios.

Therefore, the incentive for a lane change is driven by the distance between the ego-vehicle and a slower vehicle in the same lane. If this distance falls below the defined threshold and the ego-vehicle remains in the same lane, a proportional penalty is applied, more closer to the slow vehicle, more larger the penalty is. However, setting the threshold too high may trigger unnecessary or ill-timed lane changes, potentially placing the vehicle in more complex or risky situations later. On the other hand, a threshold that is too low may not allow sufficient time or space for safe maneuvering. This is especially problematic because our framework requires the ego-vehicle to consistently choose the correct lane for around 10 consecutive timesteps to complete a successful lane change, a too short safety threshold would end up leaving little room for error or recovery.

\begin{table}[htbp]
\centering
\caption{Collisions over safety thresholds.}
\begin{tabular}{{llll}}
\hline
\textbf{Safety Threshold} & \textbf{Avg.} & \textbf{Std.} & \textbf{Best}\\ 
\hline
2.0  &  6.67 & 1.64 & 2 \\
1.5   &  5.83 & 3.31 & 1  \\
\underline{1.0}  &  \underline{2.67} & \underline{3.20} & \underline{0} \\
0.5  &  5.5 & 1.52 & 3 \\
\hline
\end{tabular}
\label{tab:performance_thres}
\end{table}

\subsubsection{Aha Moment}
\begin{figure}[t]
    \centering
    \begin{subfigure}[b]{0.3\textwidth}
        \includegraphics[width=\textwidth]{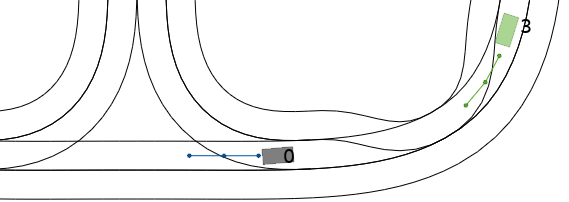}
        \caption{Timestep 1106.}
    \end{subfigure}
    
    \hfill
    
    \begin{subfigure}[b]{0.3\textwidth}
        \includegraphics[width=\textwidth]{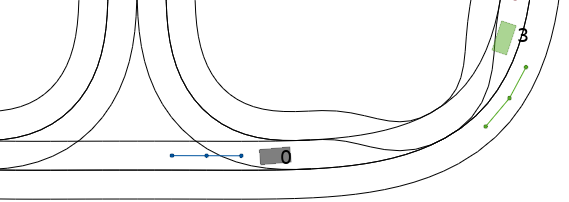}
        \caption{Timestep 1107.}
    \end{subfigure}
    
    \hfill
    
    \begin{subfigure}[b]{0.3\textwidth}
        \includegraphics[width=\textwidth]{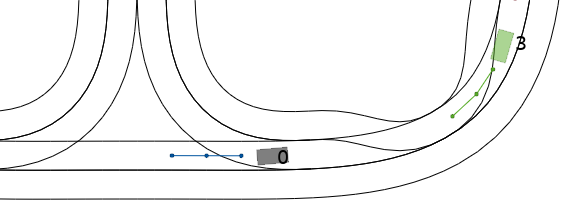}
        \caption{Timestep 1108.}
    \end{subfigure}
    
\caption{Hesitation in decision-making.}
\label{fig:hesitation}
\end{figure}

It is also exciting to share an “aha moment” discovered during our experiments, an unexpected behaviors that providing valuable insights. As shown in Figure~\ref{fig:hesitation}, during a lane-keeping scenario, the ego-vehicle (number 3, green color) briefly initiated a lane change but reverted after just one timestep. The likely explanation is that it detected a slower vehicle ahead, but the distance was still way above the defined threshold, meaning a lane change was not yet encouraged by the proposed RL framework, but it is also not prohibited though. After some hesitation, the ego-vehicle decided to remain in current lane.

As previously mentioned, a lane change is only considered successful if the ego-vehicle makes the correct decision consistently for around 10 timesteps. About unexpected moment highlights the robustness of our system, demonstrating a level of fault tolerance. As shown in Table~\ref{tab:performance_thres}, the best-performing configuration achieved nearly collision-free results. The proposed framework is capable of handling occasional false signals but still executes the lane change reliably when it becomes necessary.

\section{Conclusions}\label{sec:conclusions}

In this work, we presented a novel framework for decision-making and motion planning in CAVs, where the two modules are tightly coupled through an MPRL setup. The proposed framework demonstrates scalability, interpretability, flexibility, and theoretically high efficiency, drawing inspiration from the MoE paradigm to solve complex problems via multiple specialized small networks—the core source of its efficiency. This architecture can be readily extended to diverse learning domains. Our simulation results substantiate these claims, showing that the RL agents under this framework are able to consistently make correct, timely decisions to avoid collisions and navigate high-risk traffic scenarios.

However, several potential improvements remain to be addressed. Simulation results reveal instances of dangerously close inter-vehicle distances especially during overtaking, which, while not causing collisions, may pose real-world safety risks. We plan to incorporate our previous works on control barrier functions to enforce safety \cite{xu2025learningbased}\cite{xu2025realtime}\cite{ xu2025highorder}. However, the
introduction of CBF may adversely affect the dynamics of decision-making and thus warrants further investigation. Moreover, hesitation observed during simulation could be mitigated using CBF as well, by computing a safe decision set at each timestep to constrain the outputs of the decision-making module.


\bibliographystyle{IEEEtran}
\bibliography{00_literature}

\end{document}

%% file: 00_acronyms.tex
\DeclareAcronym{ap}{
    short = AP,
    long  = Action Propagation,
}
\DeclareAcronym{c2c}{
    short = C2C,
    long  = Center-to-Center,
}
\DeclareAcronym{cav}{
    short = CAV,
    long  = Connected and Automated Vehicle,
}
\DeclareAcronym{cbf}{
    short = CBF,
    long  = Control Barrier Function,
}
\DeclareAcronym{cg}{
    short = CG,
    long = Center of Gravity,
    short-plural = s,
    long-plural-form = Centers of Gravity,
}
\DeclareAcronym{cnn}{
    short = CNN,
    long  = Convolutional Neural Network
}
\DeclareAcronym{cpm}{
    short = CPM,
    long  = Cyber-Physical Mobility
}
\DeclareAcronym{cpmlab}{
    short = CPM Lab,
    long  = Cyber-Physical Mobility Lab
}
\DeclareAcronym{dmpc}{
    short = DMPC,
    long  = distributed model predictive control
}
\DeclareAcronym{dql}{
    short = DQL,
    long  = Deep Q-Learning
}

\DeclareAcronym{il}{
    short = IL,
    long  = Imitation Learning,
    short-indefinite = an,
}
\DeclareAcronym{mappo}{
    short = MAPPO,
    long  = Multi-Agent \ac{ppo},
    short-indefinite = an,
}
\DeclareAcronym{maddpg}{
    short = MADDPG,
    long  = Multi-Agent Deep Deterministic Policy Gradient,
    short-indefinite = an,
}
\DeclareAcronym{mas}{
    short = MAS,
    long  = Multi-Agent System,
    short-indefinite = an,
}
\DeclareAcronym{mdp}{
    short = MDP,
    long  = Markov decision process,
    short-indefinite = an,
}
\DeclareAcronym{mg}{
    short = MG,
    long  = Markov Game,
    short-indefinite = an,
}
\DeclareAcronym{ml}{
    short = ML,
    long  = Machine Learning,
    short-indefinite = an,
}
\DeclareAcronym{mtv}{
    short = MTV,
    long  = Minimum Translation Vector,
    short-indefinite = an,
}
\DeclareAcronym{mpc}{
    short = MPC,
    long  = model predictive control,
    short-indefinite = an,
}
\DeclareAcronym{marl}{
    short = MARL,
    long  = Multi-Agent Reinforcement Learning,
    short-indefinite = an,
}
\DeclareAcronym{ocp}{
    short = OCP,
    long  = Optimal Control Problem,
    short-indefinite = an,
    long-indefinite = an,
}
\DeclareAcronym{per}{
    short = PER,
    long  = Prioritized Experience Replay
}
\DeclareAcronym{pomdp}{
    short = POMDP,
    long  = Partially Observable \ac{mdp}
}
\DeclareAcronym{pomg}{
    short = POMG,
    long  = Partially Observable \ac{mg}
}
\DeclareAcronym{ppo}{
    short = PPO,
    long  = Proximal Policy Optimization
}
\DeclareAcronym{rhc}{
    short = RHC,
    long  = receding horizon control,
    short-indefinite = an,
}
\DeclareAcronym{rl}{
    short = RL,
    long  = Reinforcement Learning,
    short-indefinite = a,
}
\DeclareAcronym{sat}{
    short = SAT,
    long = Separating Axis Theorem,
    short-indefinite = an
}
\DeclareAcronym{som}{
    short = SOM,
    long  = Self Other-Modeling,
    short-indefinite = a,
}
\DeclareAcronym{zsg}{
    short = ZSG,
    long  = Zero-Shot Generalization,
}

%% file: 00_copyright.tex
\providecommand{\doi}{????/placeholder-doi} 

\makeatletter
\def\ps@IEEEtitlepagestyle{%
    \def\@oddfoot{\mycopyrightnotice}%
    \def\@evenfoot{}%
}
\makeatother

\ifpreprint
    \ifpublished
        \def\mycopyrightnotice{%
            \begin{minipage}{\textwidth}
                \centering \scriptsize
                \href{https://doi.org/\doi}{DOI: \doi} \copyright \the\year{} \copyrightOwnerFull{}. 
                Personal use of this material is permitted. Permission from \copyrightOwnerShort{} must be obtained for all other uses, in any current or future media, including reprinting/republishing this material for advertising or promotional purposes, creating new collective works, for resale or redistribution to servers or lists, or reuse of any copyrighted component of this work in other works.\hfill
            \end{minipage}
            \gdef\mycopyrightnotice{}
        }
    \else 
        \def\mycopyrightnotice{%
            \begin{minipage}{\textwidth}
                \centering \scriptsize
                This work has been submitted to the \copyrightOwnerFull{} for possible publication. Copyright may be transferred without notice, after which this version may no longer be accessible.
            \end{minipage}
            \gdef\mycopyrightnotice{}
        }
    \fi
\else
    \def\mycopyrightnotice{}
\fi